\documentclass[10pt,twocolumn,letterpaper]{article}

\usepackage{cvpr}              % To produce the CAMERA-READY version
\usepackage{xcolor}
\usepackage{graphicx}
\usepackage[normalem]{ulem}  

\usepackage{color,soul}

\usepackage{booktabs}
\usepackage{multirow}
\usepackage{colortbl}
\usepackage{amsmath}
\usepackage{bm}

\definecolor{cvprblue}{rgb}{0.21,0.49,0.74}
\usepackage[pagebackref,breaklinks,colorlinks,allcolors=cvprblue]{hyperref}

\def\paperID{}
\def\confName{CVPR}
\def\confYear{2026}

\title{\textit{4DSurf}: High-Fidelity Dynamic Scene Surface Reconstruction}

\author{
Renjie Wu$^{1}$ \quad
Hongdong Li$^{1,3}$ \quad
Jose M. Alvarez$^{2}$ \quad
Miaomiao Liu$^{1}$ \\
\vspace{0.5em}
$^{1}$Australian National University ~ $^{2}$NVIDIA~ $^{3}$Amazon\\
\vspace{0.5em}
{\tt\small \{renjie.wu, hongdong.li, miaomiao.liu\}@anu.edu.au \quad josea@nvidia.com}
}

\begin{document}
\maketitle

\begin{abstract}
This paper addresses the problem of dynamic scene surface reconstruction using Gaussian Splatting (GS), aiming to recover temporally consistent geometry. While existing GS-based dynamic surface reconstruction methods can yield superior reconstruction, they are typically limited to either a single object or objects with only small deformations, struggling to maintain temporally consistent surface reconstruction of large deformations over time. We propose ``\textit{4DSurf}'', a novel and unified framework for generic dynamic surface reconstruction that does not require specifying the number or types of objects in the scene, can handle large surface deformations and temporal inconsistency in reconstruction. The key innovation of our framework is the introduction of Gaussian deformations induced Signed Distance Function Flow Regularization that constrains the motion of Gaussians to align with the evolving surface. To handle large deformations, we introduce an Overlapping Segment Partitioning strategy that divides the sequence into overlapping segments with small deformations and incrementally passes geometric information across segments through the shared overlapping timestep. Experiments on two challenging dynamic scene datasets, Hi4D and CMU Panoptic, demonstrate that our method outperforms state-of-the-art surface reconstruction methods by 49\% and 19\% in Chamfer distance, respectively, and achieves superior temporal consistency under sparse-view settings. 
\end{abstract}

\section{Introduction}
\begin{figure}[t]
    \centering
    \includegraphics[width=\columnwidth]{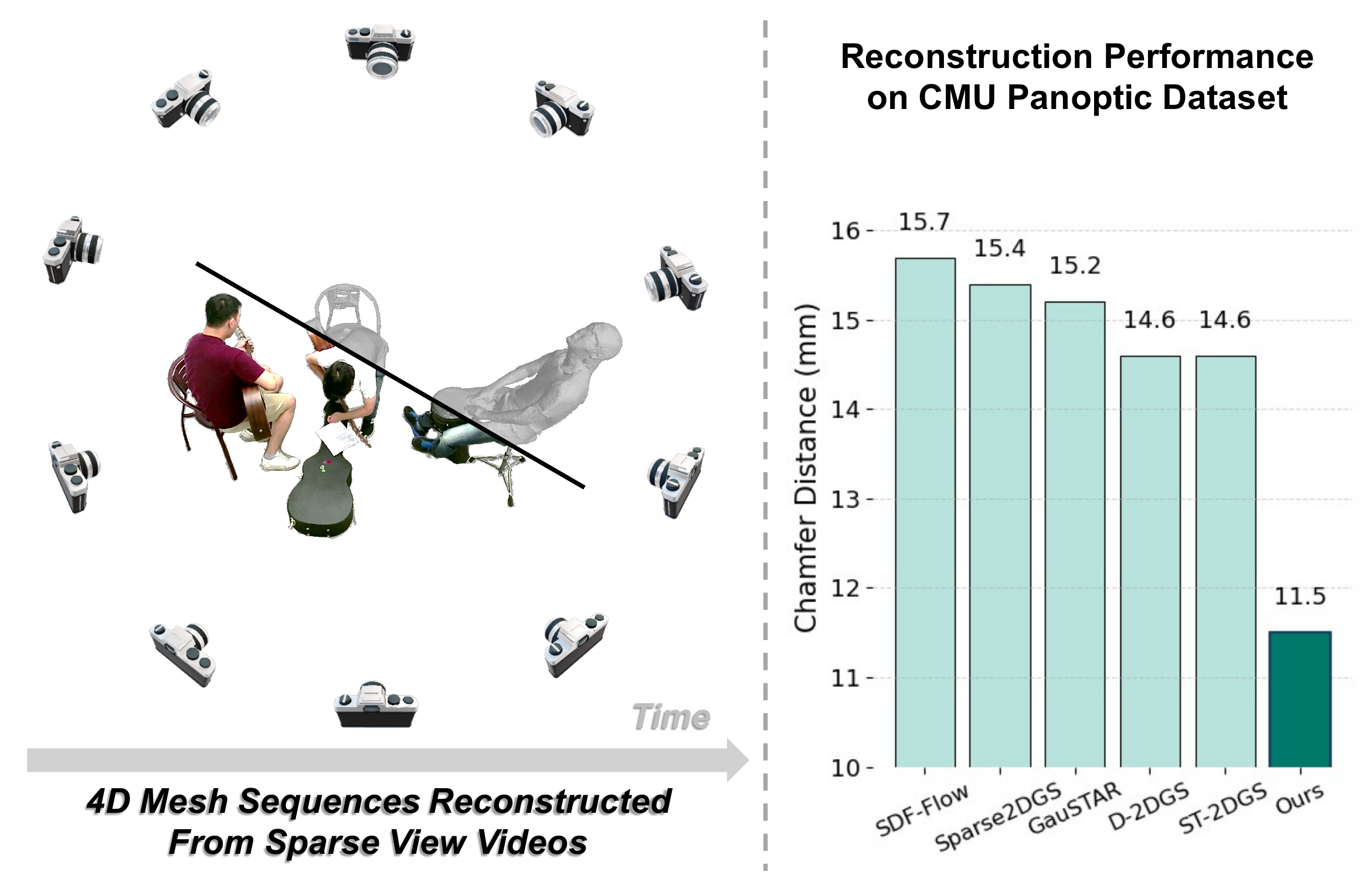}
    \caption{
    \textit{Left}: Using only a sparse set of input videos.
    \textit{Right}: Our approach sets a new state-of-the-art benchmark in surface reconstruction for dynamic scenes on the CMU Panoptic dataset~\cite{Joo_2017_TPAMI}, compared with recent dynamic surface reconstruction methods (Neural SDF-Flow~\cite{mao2024neural}, Sparse2DGS~\cite{wu2025sparse2dgs}, GauSTAR~\cite{zheng2025gaustar}, D-2DGS~\cite{zhang2024dynamic}, and ST-2DGS~\cite{wang2024space}).
    }
    \label{fig:teaser}
\end{figure}
Dynamic surface reconstruction aims to recover temporally consistent 3D geometry from video sequences, serving as a foundation for numerous applications such as digital avatars and virtual reality.~Unlike the reconstruction of static surfaces that recovers a single shape, dynamic surface reconstruction must faithfully model continuous deformations for a scene of multiple shapes over time.
Here, we address dynamic surface reconstruction from sparse-view video sequences (fewer than 10 views), a practical setup, while maintaining large scene coverage~(shown in \cref{fig:teaser}).

%%%%%%%%%%%%%
Recent advances in neural implicit and Gaussian-based representations have significantly advanced dynamic surface reconstruction. 
Dynamic Neural Radiance Fields (NeRFs)~\cite{cai2022neural, wu2023dove, biswas2024tfs, mao2024neural, wang2024morpheus} can capture detailed geometry and view-dependent effects, yet suffer from slow optimization and limited scalability. 
In contrast, Gaussian Splatting (GS)-based approaches~\cite{Wu_2024_CVPR, huang2024sc, wang2025freetimegs, wang2025monofusion} enable real-time rendering and efficient optimization, but often struggle to reconstruct accurate geometry. 
To achiever better geometry, several GS–based dynamic surface reconstruction methods~\cite{liudynamic, li2024dgns, ma2024mags, zhang2024dynamic, cai2024dynasurfgs} are proposed, but they only perform well on the case of a single object with small deformations.
Therefore, they struggle with large deformations, which often result in surface jitter and inconsistent geometry over time.
Some methods~\cite{lee2025geoavatar, jiang2024multiply, lee2024contactfield} introduce priors such as SMPL-X~\cite{pavlakos2019expressive} for human-specific dynamic surface reconstruction or depth and normal priors from pre-trained models~\cite{lee2025geoavatar} to regularize the surface reconstruction.
%%%%%%%%%%%

By contrast, we present ``\textit{4DSurf}'', a prior-free method to handle an unconstrained scene that may contain multiple (non-) rigid shapes and undergo large deformations.
Furthermore, we address the issues of surface jittering and temporal inconsistency in existing methods by 
constraining the temporal evolution of the surface. 
It is well known that a static Signed Distance Function (SDF) field can represent arbitrary surface. When extended to dynamic scenes, a time-dependent SDF field can be used to represent the surface that changes consistently over time.
Therefore, we can leverage SDF flow~\cite{mao2024neural} to characterize the temporal surface evolution, which is the temporal derivative of time-dependent SDF field and represents how signed distances evolve over the time.
The SDF flow constrains that, for any locally smooth surface point, its temporal change equals the negative projection of its scene flow onto the surface normal~\cite{mao2024neural}.
To achieve Gaussian-based temporal SDF field representation, we introduce Gaussian Velocity Field to describe the deformation of Gaussians from canonical shape to any timestep by predicting the motion of Gaussians. It explicitly models the scene flow of any surface point on each Gaussian.
We then can derive SDF flow from the motion of Gaussians.
Moreover, our dynamic Gaussian representation enables the approximation of SDF flow from rendered depth maps, providing an additional constraint from geometry changes in 3D space.
Together, these constraints promote temporally consistent surface reconstruction. 
%%%%%%%%%%%

To tackle large deformations over long sequence, we adopt an Overlapping Segment Partitioning strategy by dividing the sequence into overlapping segments trained incrementally. The geometry of each segment is modeled by a canonical shape and its associated learnable Gaussian Velocity Field. Furthermore, we develop a Low-Rank Adaptation (LoRA)~\cite{hu2022lora}-based incremental motion tuning to model the Gaussian Velocity Field for each segment, efficiently adapting motion parameters with less storage. This makes our method scalable for handling long motion sequences without losing performance, as shown in our experiments. Our main contributions are summarized as follows:
\begin{itemize}
    \item We propose a prior-free, generic geometry- and motion-consistent surface reconstruction method ``\textit{4DSurf}'' of dynamic scenes from sparse view videos.    

    \item We enforce temporally consistent reconstruction by regularizing the surface evolution so that the SDF flow matches that induced by Gaussian deformations.
    
    \item We develop an Overlapping Segment Partitioning strategy to handle large deformations, which can mitigate error accumulation and enhance scalability, and an incremental motion tuning variant that further reduces storage usage while maintaining competitive performance.
\end{itemize}
Extensive experiments on two challenging dynamic scene datasets~\cite{yin2023hi4d,Joo_2017_TPAMI} with multiple shapes and varying deformations, demonstrating our method achieves state-of-the-art performance and strong generality on reconstructing dynamic scene surface from sparse view videos.

\section{Related Work}
\noindent \textbf{Novel View Synthesis for Dynamic Scenes.} 
NeRFs~\cite{mildenhall2021nerf} have greatly advanced novel view synthesis and been extended to dynamic scenes. Dynamic NeRFs can be roughly divided into three types: (1) those accelerating training and rendering via representation decoupling or hash encoding~\cite{shao2023tensor4d, Ik2023HumanRFHN, Lin2023HighFidelityAR, Wang2023MaskedSH}; (2) those using deformation fields with a canonical space~\cite{park2021hypernerf, park2021nerfies, tretschk2021non, pumarola2021d}; and (3) incremental ones reusing a static NeRF from previous frames~\cite{song2023nerfplayer, li2022streaming}. 
Recent 3D-GS~\cite{kerbl20233d} achieves a better efficiency–quality trade-off and inspires dynamic GS. Some model dynamics via Gaussian-based canonical spaces and deformation fields~\cite{yang2024deformable, huang2024sc}, while others incorporate hash encoding or decoupled representations for faster optimization~\cite{Wu_2024_CVPR, xu2024grid4d}. Incremental schemes reuse Gaussians from prior timesteps~\cite{yan2025instant, sun20243dgstream, gao2024hicom, zheng2025gaustar}. Other works extend Gaussians into higher-dimensional space~\cite{yang2024real, gao20257dgs}, enrich Gaussian attributes~\cite{li2024spacetime, wang2025freetimegs}, or leverage priors such as optical flow or depth from foundation models~\cite{lin2024gaussian,wang2025monofusion}.
However, most of them neglect geometric representation over time, resulting in temporally inaccurate and inconsistent geometry.

\smallskip

\noindent \textbf{Dynamic Surface Reconstruction.}
Early works~\cite{kanazawa2018learning, li2020online, xue2023nsf, pan2019deep, wang2018pixel2mesh} deform predefined templates, while NeRF-based methods~\cite{cai2022neural, wu2023dove, biswas2024tfs, mao2024neural, wang2024morpheus} learn implicit fields. However, template-based methods rely heavily on templates, and NeRF-based methods suffer from slow training and limited scalability.
Recently, many GS-based dynamic surface reconstruction methods have been proposed. DG-Mesh~\cite{liudynamic} integrates deformable Gaussians with a differentiable Poisson solver~\cite{peng2021shape}, while later methods adopt 2D-GS~\cite{zhang2024dynamic, wang2024space} or planar-based Gaussians~\cite{cai2024dynasurfgs}. MaGS~\cite{ma2024mags} jointly optimizes mesh vertices and Gaussians, DGNS~\cite{li2024dgns} couples SDF-based NeRF with dynamic GS, and GauSTAR~\cite{zheng2025gaustar} adaptively regenerates Gaussians for finer surfaces. Human-specific reconstruction methods~\cite{lee2025geoavatar, jiang2024multiply, lee2024contactfield} often depend on priors such as SMPL-X~\cite{pavlakos2019expressive}.
Despite recent progress, above methods are still confined to specific scenarios (e.g., single-object or multi-human setups with dense views) and rely on priors like depth, optical flow, normals, or SMPL-X. Some recent works~\cite{mao2024neural, wang2024space} relax such foundation priors but still struggle with large deformations.
In contrast, we pursue a generic dynamic scene surface reconstruction method from sparse view videos, which is not dedicated to any specific objects and can handle large deformations.

%%%%%%%%%%%%%%%%%%%%%%%%%%%%%%%%%%%%%%%%%%%%%

\begin{figure*}[t]
    \centering
    \includegraphics[width=0.95\textwidth]{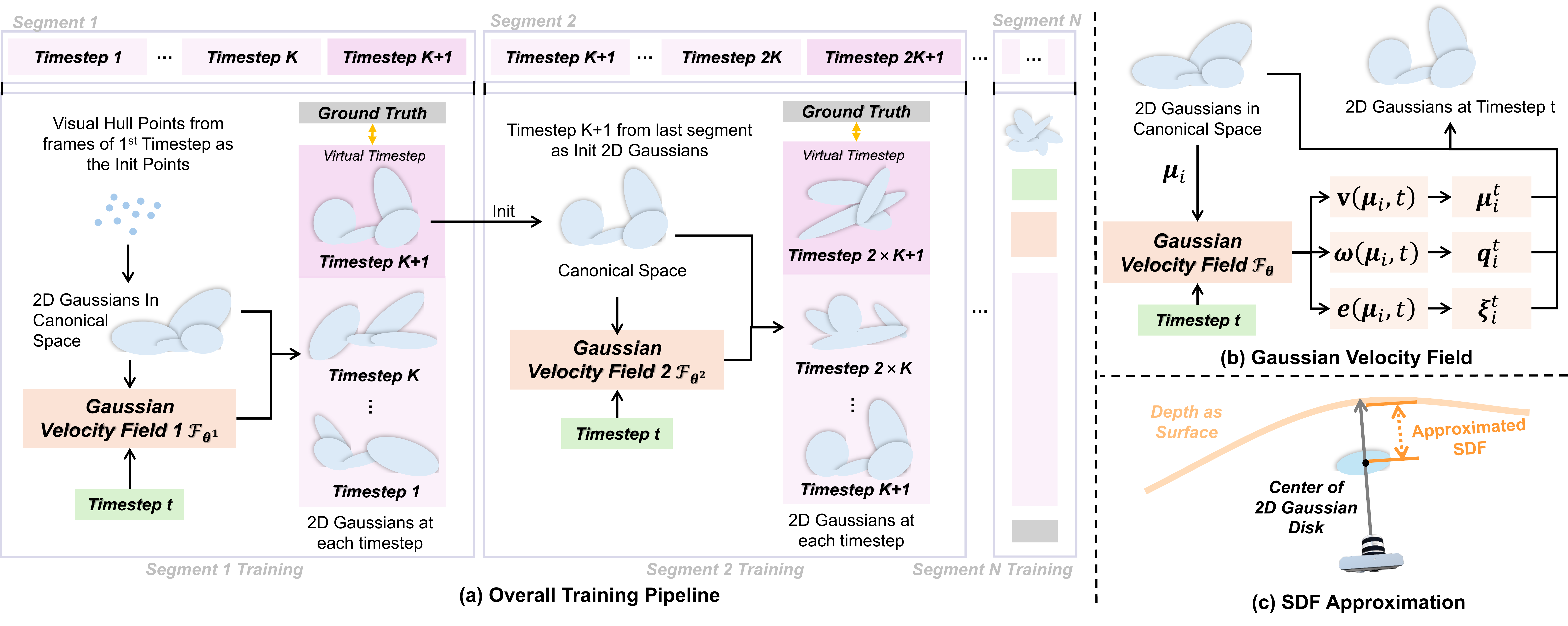}
    \caption{ 
            Overview: 
            \underline{\textbf{(a) Overall Training Pipeline.}}
            We first divide the sequence into $N$ segments, each containing $K{+}1$ timesteps with one overlapping virtual timestep. For the $1^{\text{st}}$ segment, the initialization is derived from the visual hull reconstructed from all frames of its first timestep. After training the first segment, the Gaussians of virtual timestep serves as the initialization for training the next segment. Each segment maintains its own canonical space and Gaussian Velocity Field.
            \underline{\textbf{(b) Gaussian Velocity Field.}}
            Given the Gaussian center $\bm{\mu}_i$ in the canonical space and a specific timestep $t$, the Gaussian Velocity Field $\mathcal{F}_{\bm{\theta}}(\cdot)$ predicts its velocity $\mathbf{v}(\bm{\mu}_i, t)$, angular velocity $\bm{\omega}(\bm{\mu}_i, t)$ and expansion velocity $\bm{e}(\bm{\mu}_i, t)$ at timestep $t$. These are then converted to position $\bm{\mu}_i^{t}$, rotation $\bm{q}_i^{t}$, and scale $\bm{\xi}_i^{t}$, which are fed into the differentiable rasterizer for image rendering.
            \underline{\textbf{(c) SDF Approximation.}}
            Following previous works~\cite{guedon2023sugar, newcombe2011kinectfusion}, we compute the distance between the center and its corresponding depth point to estimate the signed distance.
            }
    \label{fig:method}
\end{figure*}

%%%%%%%%%%%%%%%%%%%%%%%%%%%%%%%%%%%%%%%%%%%%%%

\section{Preliminary: 2D Gaussian Splatting}
Our goal is to model geometry, therefore we adopt 2DGS~\cite{Huang2DGS2024} in our framework, as it can model better geometry compared to 3DGS~\cite{kerbl20233d}.
Each primitive is defined by a center $\bm{\mu} \in \mathbb{R}^3$, two orthogonal tangent vectors $\mathbf{t}_u,\mathbf{t}_v$ (parameterized via quaternion $\bm{q}$), and a scale vector $\bm{\xi}=(\xi_u,\xi_v)$.  
The primitive is defined in the local tangent plane as $P(u,v)=\bm{\mu}+\xi_u \mathbf{t}_u u+\xi_v \mathbf{t}_v v$,  
and the Gaussian value can be calculated by $\mathcal{G}(\mathbf{u})=\exp(-(u^2+v^2)/2)$.  
Each primitive also carries an opacity $\alpha$ and a view-dependent color $\mathbf{c}$ represented by spherical harmonics.
The pixel color is rendered via alpha blending:
$
    \hat{\mathbf{c}}(\mathbf{p}) = \sum_{i} \mathbf{c}_i \text{w}_i,  \text{w}_i = \alpha_i \mathcal{G}_i(\mathbf{u}(\mathbf{p})) 
    \prod_{j=1}^{i-1} \left( 1 - \alpha_j \mathcal{G}_j(\mathbf{u}(\mathbf{p})) \right).
$
Depth is rendered as:
$
    \hat{D}(\mathbf{p}) = \sum_{i=1} \text{w}_i \, \mathrm{d}_i/(\sum_{i=1} \text{w}_i + \epsilon),
$
where $\mathrm{d}_i$ is the intersection depth between the ray and the $i$-th Gaussian disk.

%%%%%%%%%%%%%%%%%%%%%%%%%%%%%%%%%%%%%%%%%%%%%%%%%%%%%%%%%%%%%%%%%%%%%%%%%%%%%
\section{Method}
We present our overall training pipeline in~\cref{fig:method}(a).
In this section, we first describe our SDF Flow Regularization (\cref{sec:sdf-flow}) induced by Gaussian Velocity Field.
Then, we introduce our Overlapping Segment Partitioning strategy that can handle large deformations (\cref{sec:partitioning}).
Next, we introduce the incremental motion tuning to minimize storage while maintaining competitive performance (\cref{sec:tuning}).
Lastly, we present the overall training objective (\cref{sec:loss}).

%%%%%%%%%%%%%%%%%%%%%%%%%%%%%%%%%%%%%%%%%%%%%%%%%%%%%%%%%%%%%%%%%%%%%
\subsection{SDF Flow Regularization}
\label{sec:sdf-flow}
Our method enforces consistency between SDF flow derived from the motion of the Gaussians and SDF flow estimated from 3D geometric changes, jointly yielding temporally consistent surface evolution.
Below we first revisit the definition of SDF flow in~\cite{mao2024neural} followed by our defined Gaussian Velocity Field and the derived regularization.

\subsubsection*{Revisit SDF Flow from Point-Wise Motion.}
SDF field for static scene defines a scalar field $ s(\mathbf{\hat{x}}): \mathbb{R}^3 \to \mathbb{R},$ where $s(\mathbf{\hat{x}})$ denotes the signed distance from point $\mathbf{\hat{x}}$ to the closest surface. The surface is given by the zero-level set $\{\mathbf{\hat{x}}\mid s(\mathbf{\hat{x}})=0\}$, and the normal is defined as $\mathbf{n}(\mathbf{\hat{x}})=\nabla_{\mathbf{\hat{x}}} s(\mathbf{\hat{x}})$. 
When extended to dynamic scenes, the SDF field becomes time-dependent, $s(\mathbf{\hat{x}},t):\mathbb{R}^3\times\mathbb{R}\to\mathbb{R}$,
and its temporal derivative $\tfrac{\partial s}{\partial t}$ characterizes how the signed distance evolves.
Following~\cite{mao2024neural}, when $\Delta t \to 0$, 
for a locally rigid point with line velocity $\mathbf{v}$ and angular velocity $\boldsymbol{\omega}$,
its SDF flow is given by:
\begin{equation}
\frac{\partial s}{\partial t}
= \lim_{\Delta t\to 0}\frac{\Delta s}{\Delta t}
=-\frac{\partial \mathbf{\hat{x}}}{\partial t}^{\!\top}\mathbf{n}(\mathbf{\hat{x}})= -(\boldsymbol{\omega}\times\mathbf{\hat{x}} + \mathbf{v})^{\!\top}\mathbf{n}(\mathbf{\hat{x}}),
\label{eq:sdf-flow-general}
\end{equation}
which directly links SDF changes to the underlying scene flow $\frac{\partial \mathbf{\hat{x}}}{\partial t}$, demonstrating how the surface evolves over time.

However, Eq.~\eqref{eq:sdf-flow-general} only considers point-wise motion independently on the surface and its relations to its SDF changes, which cannot be directly applied for the points on the Gaussians in the canonical space.
Therefore, we introduce the Gaussian Velocity Field to describe the deformations of Gaussians from canonical shape to any timestep by modeling the motion of Gaussians. 
With such Gaussian Velocity Field, we can explicitly model the scene flow of any surface point on each Gaussian.
Then we can derive the relationship between the motion of the Gaussian and its induced SDF changes to all surface points of the Gaussian.
In the following, we first describe the Gaussian Velocity Field and then derive the SDF flow from the motion of Gaussians. 

\subsubsection*{Gaussian Velocity Field}
\label{sec:field}
To explicitly describe the scene flow of Gaussians, we use Gaussian Velocity Field to predict motions of Gaussians to model deformations at each timestep instead of directly predicting the absolute deformation.
The \cref{fig:method}(b) shows the overall procedure.
Specifically, given the center $\bm{\mu}_i$ of the $i^{\text{th}}$ Gaussian in the canonical space and the timestep $t \in \mathbb{R}$, the Gaussian Velocity Field $\mathcal{F}_{\bm{\theta}}$ predicts three types of motion parameters: 
velocity $\mathbf{v}(\bm{\mu}_i, t) \in \mathbb{R}^3$, angular velocity $\bm{\omega}(\bm{\mu}_i, t) \in \mathbb{R}^3$, and expansion velocity $\bm{e}(\bm{\mu}_i, t) \in \mathbb{R}^2$ (velocity of scale changes). 
Formally,
$\mathbf{v}(\bm{\mu}_i, t), \; {\bm{\omega}}(\bm{\mu}_i, t), \; \bm{e}(\bm{\mu}_i, t) = \mathcal{F}_{\bm{\theta}}\big(\gamma(\bm{\mu}_i), \gamma(t)\big)$, 
where $\gamma(\cdot)$ denotes the positional encoding function. 
Then, we can obtain the following parameters describing the deformation of the $i^{\text{th}}$ Gaussian at timestep $t$:
$
    \bm{\mu}_i^{t} = \bm{\mu}_i + \mathbf{v}(\bm{\mu}_i, t)\, t, 
    \bm{q}_i^{t} = \phi\big({\boldsymbol \omega}(\bm{\mu}_i, t)\, t\big) \otimes \bm{q}_i, 
    \bm{\xi}_i^{t} = \bm{\xi}_i + \bm{e}(\bm{\mu}_i, t)\, t,
$
where $\phi(\cdot)$ represents a function that can convert rotation vectors to quaternions, and $\otimes$ denotes quaternion multiplication. 
These Gaussian parameters can enable differentiable rasterization rendering at arbitrary timestep within each segment.

\subsubsection*{SDF Flow from Motion of Gaussians}
\label{sec:ana-sdf-flow}
Given the definition of above Gaussian Velocity Field, we then can derive the SDF flow induced by the motion of Gaussians, which can be introduced as a constraint for ensuring temporally consistent reconstruction of surfaces. 

\noindent\textbf{Assumption.}
\textit{
At timestep $t$, the motion of a 3D point $\mathbf{x}\in\mathbb{R}^3$ in the canonical space of a Gaussian can be approximated by a rigid transformation parameterized by rotation $\mathrm{R}^t\in SO(3)$ and translation $\mathrm{T}^t\in\mathbb{R}^3$. 
After an interval $\Delta t \to 0$, we can get the displacement of the point: }
\begin{equation}
    \mathbf{x}^{t+\Delta t} - \mathbf{x}^{t} = \Delta\mathrm{R}\,\mathrm{R}^t \mathbf{x} + \Delta\mathrm{T},
\end{equation}
\textit{where $\Delta\mathrm{R}\in SO(3)$ and $\Delta\mathrm{T}\in\mathbb{R}^3$ denote the incremental rotation and translation.}

\noindent\textbf{Theorem.}
\textit{
For any 3D point $\mathbf{x}$ in the canonical space of a Gaussian, 
the temporal derivative $\mathbf{f}$ of its SDF equals the negative projection of the induced scene flow onto the surface normal $\mathbf{n}$ (normal of the Gaussian):}
\begin{align}
    \mathbf{f}
    = -({\boldsymbol \omega}\times \mathrm{R}^t{\bf x} + \mathbf{v})^{\top}\mathbf{n}(\mathrm{R}^t{\bf x}),
    \label{eq:sdf-flow-gaussian}
\end{align}
\textit{where $\mathbf{f}\in\mathbb{R}$ denotes the SDF flow, 
$\boldsymbol{\omega}\in\mathbb{R}^3$ and $\mathbf{v}\in\mathbb{R}^3$ are the angular and linear velocities of the points on the Gaussian, 
and $\mathbf{n}(\mathrm{R}^t\mathbf{x})$ is the surface normal at $\mathrm{R}^t\mathbf{x}$.}

\noindent Please refer to Supplementary Material Sec. 7 for a detailed derivation.
Note that, we compute the SDF flow at center of each Gaussian in practice for efficient regularization.

\subsubsection*{SDF Flow from Geometry Changes} 
\label{sec:geo-sdf-flow}
On the other hand, SDF flow can be derived by measuring the changes of SDF values of each point in the 3D space.
The rendered depth can be interpreted as a pseudo-surface of a time-dependent SDF.
Here we look at the Gaussian centers at time $t$ and use these signed distances as our approximated SDF values, for convenient computation and compatibility with the existing GS rasterizer.
Indeed, ideally, the SDF values of Gaussian centers are zero for time $t$. However, SDF flow measures their changes due to the evolving of the surface induced by the motion of Gaussians.
We follow previous works~\cite{guedon2023sugar, newcombe2011kinectfusion} to do the SDF approximation efficiently,
which also leverages depth map as a pseudo-surface for SDF estimation.
We exploit the temporal derivative of those SDF values.
Thus, given a Gaussian center $\bm{\mu}_i^{t}$,
we can approximate the SDF value $\tilde{s}(\bm{\mu}_i^{t}, t)$ at timestep $t$ as:
\begin{equation}
\tilde{s}(\bm{\mu}_i^{t}, t) = \hat{D}(\mathbf{p}^{*},t) - d(\bm{\mu}_i^{t},t),
\label{eq:app-sdf}
\end{equation}
where $d(\bm{\mu}_i^{t},t) \in \mathbb{R}$ denotes the distance from the camera origin to the $\bm{\mu}_i^{t}$ along the optical axis, 
and $\hat{D}(\mathbf{p}^{*},t)$ represents the corresponding surface depth point at the projected pixel $\mathbf{p}^{*}$ on the depth map.
The process can be referred to \cref{fig:method}(c).
Then we can get its temporal derivative to obtain the SDF flow $\tilde{\mathbf{f}}_i \in \mathbb{R}$ from geometry changes:
\begin{equation}
    \tilde{\mathbf{f}}^{t}_i = \frac{\partial \tilde{s}(\bm{\mu}_i^{t}, t)}{\partial t} = \frac{\partial \hat{D}(\mathbf{p}^{*},t)}{\partial t} - \frac{\partial d(\bm{\mu}_i^{t},t) }{\partial t}.
    \label{eq:app-sdf-flow}
\end{equation}
See Supplementary Material Sec. 8 for more details of above formula.
By matching SDF flow from the motion of Gaussians and geometry changes, we can achieve temporally consistent surface evolution.
Formally, SDF flow regularization
$
\mathcal{L}_{\text{flow}} = 
\sum_{i} | \mathbf{f}^{t}_i - \tilde{\mathbf{f}}^{t}_i |,
$
where $\mathbf{f}^{t}_i$ and $\tilde{\mathbf{f}}^{t}_i$ denote the SDF flow from the motion of Gaussians and geometry changes of the $i^{\text{th}}$ Gaussian at timestep $t$, respectively.

\subsection{Segment Partition}
\label{sec:partitioning}
Existing works~\cite{zhang2024dynamic, li2024dgns, ma2024mags} can reconstruct dynamic scenes with small deformations using one deformation field and canonical shape. However, they struggle with large deformations.
To address this issue, we subdivide the video sequence into consecutive segments train the the model incrementally.  
We allow overlaps between neighboring segments, where each segment shares a virtual timestep with the following one (the first timestep of the next segment). Thus, each segment contains $K{+}1$ timesteps.
We name this strategy as Overlapping Segment Partitioning.
Within each segment, we model the geometry using a canonical shape along with a Gaussian Velocity Field that describes the deformation. After training one segment, the Gaussians at its virtual timestep are used as the initialization for training the next segment, ensuring consistent geometry can pass to later segments.
Also, new Gaussians will be created in previously unseen regions.
This process is shown in \cref{fig:method}(a). 
During mesh extraction, the reconstructed mesh of the virtual timestep is discarded to avoid duplication.

%%%%%%%%%%%%%%%%%%%%%%%%%%%%%%%%%%%%%%%%%%%%%%%%%%%%%%%%%%%%%%%%%%%%%

\begin{figure}[t]
    \centering
    \includegraphics[width=\columnwidth]{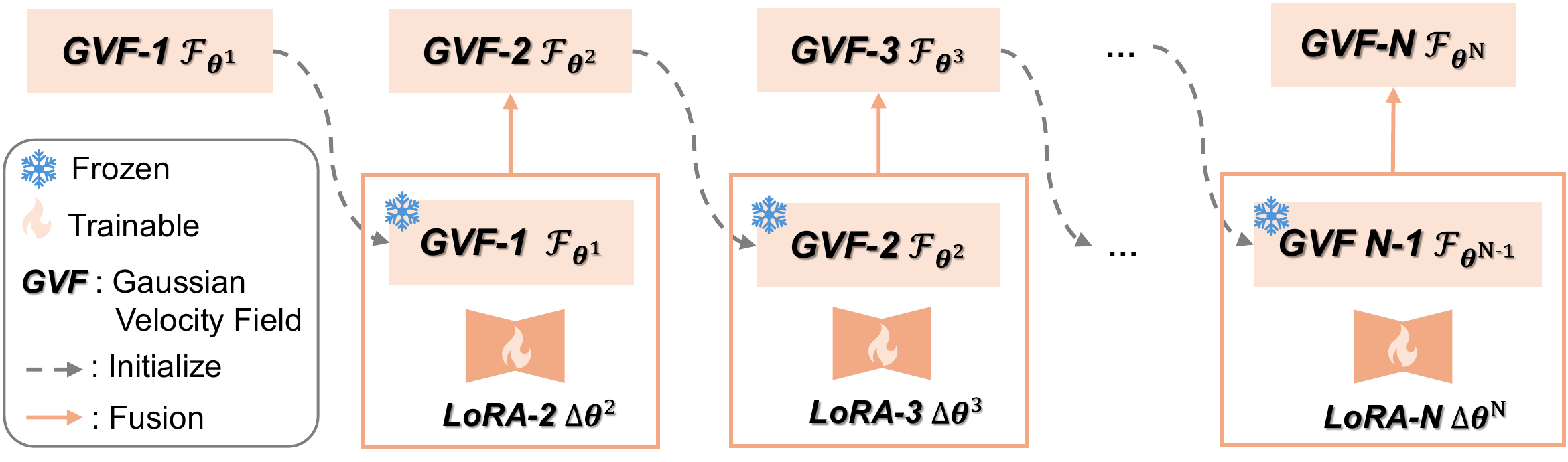}
    \caption{Incremental Motion Tuning (IMT). 
    After training the Gaussian Velocity Field of the $1^{\text{st}}$ segment, for the later $N^{\text{th}}$ segment ($N \ge 2$), its Gaussian Velocity Field $\mathcal{F}_{\bm{\theta}^{N}}$ is initialized from the $\bm{\theta}^{N-1}$ of previous segment and fine-tuned using LoRA $\Delta\bm{\theta}^{N}$.
    }
    \label{fig:lora}
\end{figure}

%%%%%%%%%%%%%%%%%%%%%%%%%%%%%%%%%%%%%%%%%%%%%%%%%%%%%%%%%%%

%%%%%%%%%%%%%%%%%%%%%%%%%%%%%%%%%%%%%%%%%%%%%%%%%%%%%%%%%%%%%%%%%%%%%

\subsection{Incremental Motion Tuning}
\label{sec:tuning}
The Overlapping Segment Partitioning-based incremental training strategy can mitigate error accumulation and handle large deformations, but it increases the number of Gaussian velocity fields and canonical shapes, leading to higher storage costs.
Merging canonical shapes of all segments is non-trivial, so that we aim to reduce storage of Gaussian Velocity Fields. 
Since all segments depict the same dynamic scene, their primary difference lies in motion. 
We propose Incremental Motion Tuning (IMT), which incrementally adapts the Gaussian Velocity Field from the previous segment in a parameter-efficient manner. 
To be specific, after training the Gaussian Velocity Field of the first segment, for later $N^{\text{th}}$ segment ($N \geq 2$), we fine-tune the Gaussian Velocity Field from its previous segment by using LoRA~\cite{hu2022lora} instead of learning a new one from scratch (see \cref{fig:lora}). 
Formally, the Gaussian Velocity Field parameters of the $N^{\text{th}}$ ($N \geq 2$) segment is defined as:
$
\bm{\theta}^{N} = \bm{\theta}^{N-1} + \Delta \bm{\theta}^{N},
$
where $\bm{\theta}^{N}$ denotes the Gaussian Velocity Field parameters of segment $N$, and $\Delta \bm{\theta}^{N} = \mathbf{A}^{N}\mathbf{B}^{N}$ represents the low-rank update, 
with $\mathbf{A}^{N} \in \mathbb{R}^{\mathbf{d} \times \mathbf{r}}$ and $\mathbf{B}^{N} \in \mathbb{R}^{\mathbf{r} \times \mathbf{d}}$, where $\mathbf{r} \ll \mathbf{d}$.
By storing only $\Delta \bm{\theta}^{N}$ for each segment, IMT can reduce the storage cost while maintain competitive performance.

%%%%%%%%%%%%%%%%%%%%%%%%%%%%%%%%%%%%%%%%%%%%%%%%%%%%%%%%%%%%%%%%%%%%%

\subsection{Optimization}
\label{sec:loss}
We employ a photometric loss $\mathcal{L}_{\text{img}}$, three regularization losses, and a mask loss $\mathcal{L}_{\text{m}}$. 
$\mathcal{L}_{\text{img}}$ follows prior works~\cite{zhang2024dynamic, huang2024sc} and combines an $\mathcal{L}_1$ with D-SSIM term. 
We also adopt the normal alignment loss $\mathcal{L}_{\text{n}}$ and depth distortion loss $\mathcal{L}_{\text{d}}$ from 2DGS~\cite{Huang2DGS2024}, which align splat normals with depth-derived normals for coherent surface and regularize geometry by encouraging concentrated ray–splat intersections to reduce depth ambiguity.
Then, followed by our SDF flow regularization $\mathcal{L}_{\text{flow}}$.
Finally, we incorporate a mask loss to reduce background artifacts. Formally, 
$\mathcal{L}_{\text{m}} = \mathcal{L}_1(\mathbf{M}^{*}, \mathbf{M})$, 
where $\mathbf{M}^{*}$ and $\mathbf{M}$ denote rendered and ground-truth alpha masks.
Here is the total training objective $\mathcal{L}_{\text{total}}$:
\begin{equation}
    \mathcal{L}_{\text{total}} = 
    \mathcal{L}_{\text{img}} + 
    \lambda_1 \mathcal{L}_{\text{n}} + 
    \lambda_2 \mathcal{L}_{\text{d}} + 
    \lambda_3 \mathcal{L}_{\text{flow}} + 
    \lambda_4 \mathcal{L}_{\text{m}},
    \label{eq:total-loss}
\end{equation}
where $\lambda_1, \lambda_2, \lambda_3, \lambda_4$ denote the weights of each loss term.

% \usepackage{booktabs}
% \usepackage{multirow}
% \usepackage{graphicx}
\begin{table*}[t]
\centering
\caption{CMU Panoptic~\cite{Joo_2017_TPAMI} comparisons. We evaluate performance with Chamfer Distance (unit: mm).
The top three results for each metric are highlighted with \colorbox[HTML]{FE996B}{\phantom{---}}, \colorbox[HTML]{FFCE93}{\phantom{---}}, and \colorbox[HTML]{FFFFC7}{\phantom{---}}, respectively. 
Ours consistently achieves the best performance on the Overall metric.
}
\label{tab:cmu}
\resizebox{0.73\textwidth}{!}{%
\begin{tabular}{@{}ccccccccccccc@{}}
\toprule
\multirow{2}{*}{Methods} &
  \multicolumn{3}{c}{\texttt{Band1}} &
  \multicolumn{3}{c}{\texttt{Ian3}} &
  \multicolumn{3}{c}{\texttt{Haggling b2}} &
  \multicolumn{3}{c}{\texttt{Pizza1}} \\
 &
  Acc $\downarrow$ &
  Comp $\downarrow$ &
  Overall $\downarrow$ &
  Acc $\downarrow$ &
  Comp $\downarrow$ &
  Overall $\downarrow$ &
  Acc $\downarrow$ &
  Comp $\downarrow$ &
  Overall $\downarrow$ &
  Acc $\downarrow$ &
  Comp $\downarrow$ &
  Overall $\downarrow$ \\ \midrule
NDR~\cite{cai2022neural} &
  15.9 &
  23.7 &
  19.8 &
  21.8 &
  20.7 &
  21.3 &
  12.5 &
  22.8 &
  17.7 &
  17.7 &
  25.0 &
  21.3 \\
Tensor4D~\cite{shao2023tensor4d} &
  17.1 &
  29.2 &
  23.2 &
  15.4 &
  22.8 &
  19.1 &
  13.7 &
  25.3 &
  19.5 &
  18.3 &
  23.5 &
  22.9 \\
Neural SDF-Flow~\cite{mao2024neural} &
  13.0 &
  21.4 &
  17.2 &
  14.1 &
  17.5 &
  15.8 &
  \cellcolor[HTML]{FE996B}8.3 &
  18.6 &
  \cellcolor[HTML]{FFFFC7}13.5 &
  11.5 &
  20.6 &
  16.1 \\
4DGS~\cite{Wu_2024_CVPR} &
  12.4 &
  22.4 &
  17.3 &
  8.8 &
  17.2 &
  13.0 &
  9.0 &
  19.9 &
  14.4 &
  12.1 &
  22.1 &
  17.1 \\
SC-GS~\cite{huang2024sc} &
  12.2 &
  22.2 &
  17.2 &
  8.4 &
  17.3 &
  12.8 &
  \cellcolor[HTML]{FE996B}8.3 &
  19.2 &
  13.8 &
  11.7 &
  22.1 &
  16.9 \\
FreeTimeGS~\cite{wang2025freetimegs} &
  12.8 &
  22.9 &
  17.9 &
  8.8 &
  17.3 &
  13.1 &
  10.6 &
  19.9 &
  15.3 &
  12.9 &
  22.5 &
  17.7 \\
Sparse2DGS~\cite{wu2025sparse2dgs} &
  12.6 &
  22.7 &
  17.7 &
  \cellcolor[HTML]{FE996B}7.8 &
  17.2 &
  \cellcolor[HTML]{FFFFC7}12.5 &
  \cellcolor[HTML]{FFCE93}8.4 &
  20.3 &
  14.4 &
  \cellcolor[HTML]{FFFFC7}11.1 &
  22.7 &
  16.9 \\
Space-Time-2DGS~\cite{wang2024space} &
  \cellcolor[HTML]{FFFFC7}11.9 &
  20.9 &
  16.4 &
  8.6 &
  \cellcolor[HTML]{FFFFC7}16.5 &
  12.6 &
  8.6 &
  18.9 &
  13.7 &
  11.2 &
  20.4 &
  15.8 \\
GauSTAR~\cite{zheng2025gaustar} &
  14.2 &
  21.1 &
  17.6 &
  10.1 &
  17.3 &
  13.7 &
  11.0 &
  18.7 &
  14.8 &
  11.4 &
  \cellcolor[HTML]{FFFFC7}17.9 &
  \cellcolor[HTML]{FFFFC7}14.7 \\
Dynamic-2DGS~\cite{zhang2024dynamic} &
  12.1 &
  \cellcolor[HTML]{FFFFC7}19.9 &
  \cellcolor[HTML]{FFFFC7}16.0 &
  9.5 &
  \cellcolor[HTML]{FFCE93}15.4 &
  \cellcolor[HTML]{FFFFC7}12.5 &
  9.3 &
  \cellcolor[HTML]{FFFFC7}18.0 &
  13.7 &
  12.6 &
  19.8 &
  16.2 \\ \midrule
Ours w IMT-64 &
  \cellcolor[HTML]{FE996B}11.0 &
  \cellcolor[HTML]{FFCE93}14.5 &
  \cellcolor[HTML]{FFCE93}12.8 &
  \cellcolor[HTML]{FFCE93}8.2 &
  \cellcolor[HTML]{FE996B}12.6 &
  \cellcolor[HTML]{FE996B}10.4 &
  \cellcolor[HTML]{FFFFC7}8.5 &
  \cellcolor[HTML]{FFCE93}13.5 &
  \cellcolor[HTML]{FFCE93}11.0 &
  \cellcolor[HTML]{FE996B}10.8 &
  \cellcolor[HTML]{FFCE93}13.5 &
  \cellcolor[HTML]{FE996B}12.1 \\
Ours wo IMT &
  \cellcolor[HTML]{FFCE93}11.1 &
  \cellcolor[HTML]{FE996B}14.4 &
  \cellcolor[HTML]{FE996B}12.7 &
  \cellcolor[HTML]{FFFFC7}8.3 &
  \cellcolor[HTML]{FE996B}12.6 &
  \cellcolor[HTML]{FFCE93}10.5 &
  \cellcolor[HTML]{FFCE93}8.4 &
  \cellcolor[HTML]{FE996B}13.3 &
  \cellcolor[HTML]{FE996B}10.8 &
  \cellcolor[HTML]{FFCE93}10.9 &
  \cellcolor[HTML]{FE996B}13.4 &
  \cellcolor[HTML]{FFCE93}12.2 \\ \bottomrule
\end{tabular}%
}
\end{table*}

% \usepackage{booktabs}
% \usepackage{multirow}
% \usepackage{graphicx}
\begin{table*}[t]
\centering
\caption{Hi4D~\cite{yin2023hi4d} comparisons. 
We evaluate performance using Chamfer Distance (unit: cm). 
The top three results for each metric are highlighted in \colorbox[HTML]{FE996B}{\phantom{---}}, \colorbox[HTML]{FFCE93}{\phantom{---}}, and \colorbox[HTML]{FFFFC7}{\phantom{---}}, respectively. 
Ours significantly outperforms all baselines on the Overall metric.
}
\label{tab:hi4d}
\resizebox{\textwidth}{!}{%
\begin{tabular}{@{}ccccccccccccccccccc@{}}
\toprule
 &
  \multicolumn{3}{c}{\texttt{Cheers37}} &
  \multicolumn{3}{c}{\texttt{Talk22}} &
  \multicolumn{3}{c}{\texttt{Football18}} &
  \multicolumn{3}{c}{\texttt{Fight17}} &
  \multicolumn{3}{c}{\texttt{Basketball13}} &
  \multicolumn{3}{c}{\texttt{Backhug02}} \\
\multirow{-2}{*}{Methods} &
  Acc $\downarrow$ &
  Comp $\downarrow$ &
  Overall $\downarrow$ &
  Acc $\downarrow$ &
  Comp $\downarrow$ &
  Overall $\downarrow$ &
  Acc $\downarrow$ &
  Comp $\downarrow$ &
  Overall $\downarrow$ &
  Acc $\downarrow$ &
  Comp $\downarrow$ &
  Overall $\downarrow$ &
  Acc $\downarrow$ &
  Comp $\downarrow$ &
  Overall $\downarrow$ &
  Acc $\downarrow$ &
  Comp $\downarrow$ &
  Overall $\downarrow$ \\ \midrule
4DGS~\cite{Wu_2024_CVPR} &
  3.05 &
  1.43 &
  2.24 &
  3.73 &
  3.75 &
  3.74 &
  3.05 &
  2.22 &
  2.63 &
  2.53 &
  2.94 &
  2.73 &
  2.96 &
  2.53 &
  2.74 &
  3.05 &
  5.36 &
  4.21 \\
SC-GS~\cite{huang2024sc} &
  \cellcolor[HTML]{FFFFC7}1.58 &
  \cellcolor[HTML]{FFFFC7}0.97 &
  \cellcolor[HTML]{FFFFC7}1.27 &
  \cellcolor[HTML]{FFFFC7}1.62 &
  3.21 &
  2.41 &
  \cellcolor[HTML]{FFFFC7}1.47 &
  1.81 &
  1.64 &
  1.58 &
  2.65 &
  2.11 &
  \cellcolor[HTML]{FFFFC7}1.72 &
  1.90 &
  \cellcolor[HTML]{FFFFC7}1.81 &
  1.87 &
  5.16 &
  3.51 \\
FreeTimeGS~\cite{wang2025freetimegs} &
  2.34 &
  1.22 &
  1.78 &
  3.21 &
  2.21 &
  2.71 &
  2.20 &
  1.64 &
  1.92 &
  3.71 &
  2.14 &
  2.93 &
  3.04 &
  3.10 &
  3.07 &
  2.61 &
  3.81 &
  3.21 \\
Sparse2DGS~\cite{wu2025sparse2dgs} &
  2.66 &
  0.96 &
  1.81 &
  3.21 &
  1.86 &
  2.54 &
  1.94 &
  0.98 &
  1.46 &
  \cellcolor[HTML]{FFFFC7}1.27 &
  \cellcolor[HTML]{FFFFC7}0.75 &
  \cellcolor[HTML]{FFFFC7}1.01 &
  4.87 &
  2.10 &
  3.48 &
  2.00 &
  1.59 &
  1.79 \\
GauSTAR~\cite{zheng2025gaustar} &
  2.10 &
  2.52 &
  2.31 &
  2.74 &
  1.96 &
  \cellcolor[HTML]{FFFFC7}2.35 &
  2.74 &
  0.49 &
  1.62 &
  3.32 &
  2.25 &
  2.79 &
  4.15 &
  1.04 &
  2.59 &
  2.18 &
  1.99 &
  2.09 \\
Dynamic-2DGS~\cite{zhang2024dynamic} &
  2.62 &
  1.96 &
  2.29 &
  2.75 &
  \cellcolor[HTML]{FFFFC7}0.88 &
  \cellcolor[HTML]{FFCE93}1.82 &
  1.76 &
  \cellcolor[HTML]{FFFFC7}0.44 &
  \cellcolor[HTML]{FFFFC7}1.10 &
  2.33 &
  1.20 &
  1.77 &
  3.55 &
  \cellcolor[HTML]{FFFFC7}0.98 &
  2.27 &
  \cellcolor[HTML]{FFFFC7}1.59 &
  \cellcolor[HTML]{FE996B}0.55 &
  \cellcolor[HTML]{FFFFC7}1.07 \\ \midrule
Ours w IMT-64 &
  \cellcolor[HTML]{FFCE93}0.71 &
  \cellcolor[HTML]{FFCE93}0.25 &
  \cellcolor[HTML]{FFCE93}0.48 &
  \cellcolor[HTML]{FFCE93}1.09 &
  \cellcolor[HTML]{FE996B}0.59 &
  \cellcolor[HTML]{FE996B}0.84 &
  \cellcolor[HTML]{FFCE93}0.81 &
  \cellcolor[HTML]{FFCE93}0.30 &
  \cellcolor[HTML]{FFCE93}0.56 &
  \cellcolor[HTML]{FE996B}0.82 &
  \cellcolor[HTML]{FFCE93}0.46 &
  \cellcolor[HTML]{FFCE93}0.64 &
  \cellcolor[HTML]{FFCE93}1.09 &
  \cellcolor[HTML]{FFCE93}0.69 &
  \cellcolor[HTML]{FFCE93}0.89 &
  \cellcolor[HTML]{FE996B}0.93 &
  \cellcolor[HTML]{FFCE93}0.60 &
  \cellcolor[HTML]{FE996B}0.76 \\
Ours wo IMT &
  \cellcolor[HTML]{FE996B}0.70 &
  \cellcolor[HTML]{FE996B}0.24 &
  \cellcolor[HTML]{FE996B}0.47 &
  \cellcolor[HTML]{FE996B}1.06 &
  \cellcolor[HTML]{FFCE93}0.62 &
  \cellcolor[HTML]{FE996B}0.84 &
  \cellcolor[HTML]{FE996B}0.73 &
  \cellcolor[HTML]{FE996B}0.29 &
  \cellcolor[HTML]{FE996B}0.51 &
  \cellcolor[HTML]{FFCE93}0.91 &
  \cellcolor[HTML]{FE996B}0.35 &
  \cellcolor[HTML]{FE996B}0.63 &
  \cellcolor[HTML]{FE996B}0.90 &
  \cellcolor[HTML]{FE996B}0.50 &
  \cellcolor[HTML]{FE996B}0.70 &
  \cellcolor[HTML]{FFCE93}1.01 &
  \cellcolor[HTML]{FFFFC7}0.68 &
  \cellcolor[HTML]{FFCE93}0.84 \\ \bottomrule
\end{tabular}%
}
\end{table*}

\begin{figure*}[t]
    \centering
    \includegraphics[width=0.88\textwidth]{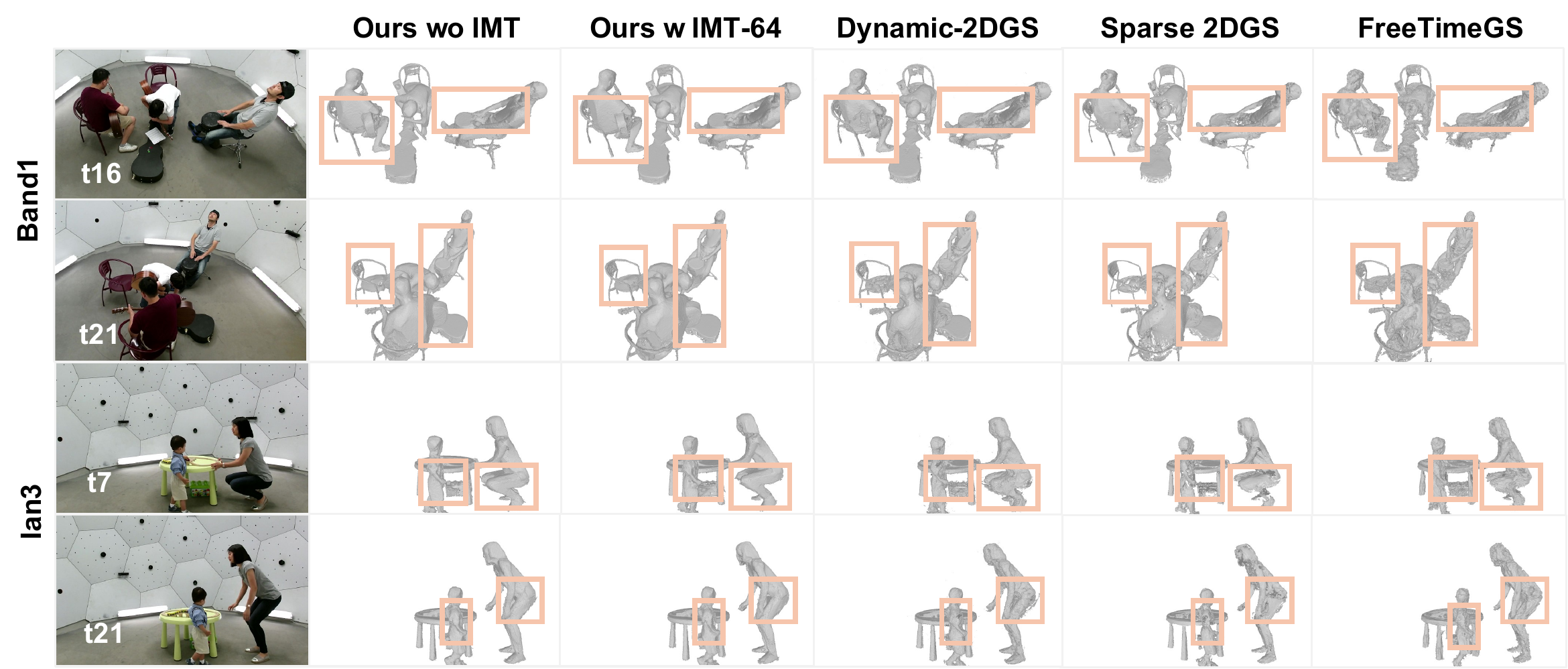}
    \caption{Qualitative results on CMU Panoptic~\cite{Joo_2017_TPAMI}. We compare our methods with three baselines (Dynamic-2DGS~\cite{zhang2024dynamic}, Sparse2DGS~\cite{wu2025sparse2dgs}, FreeTimeGS~\cite{wang2025freetimegs}) at two timesteps of the \texttt{Band1} and \texttt{Ian3} scene. Bounding boxes highlight major differences.
    }
    \label{fig:cmu}
\end{figure*}

\begin{figure*}[t]
    \centering
    \includegraphics[width=0.82\textwidth]{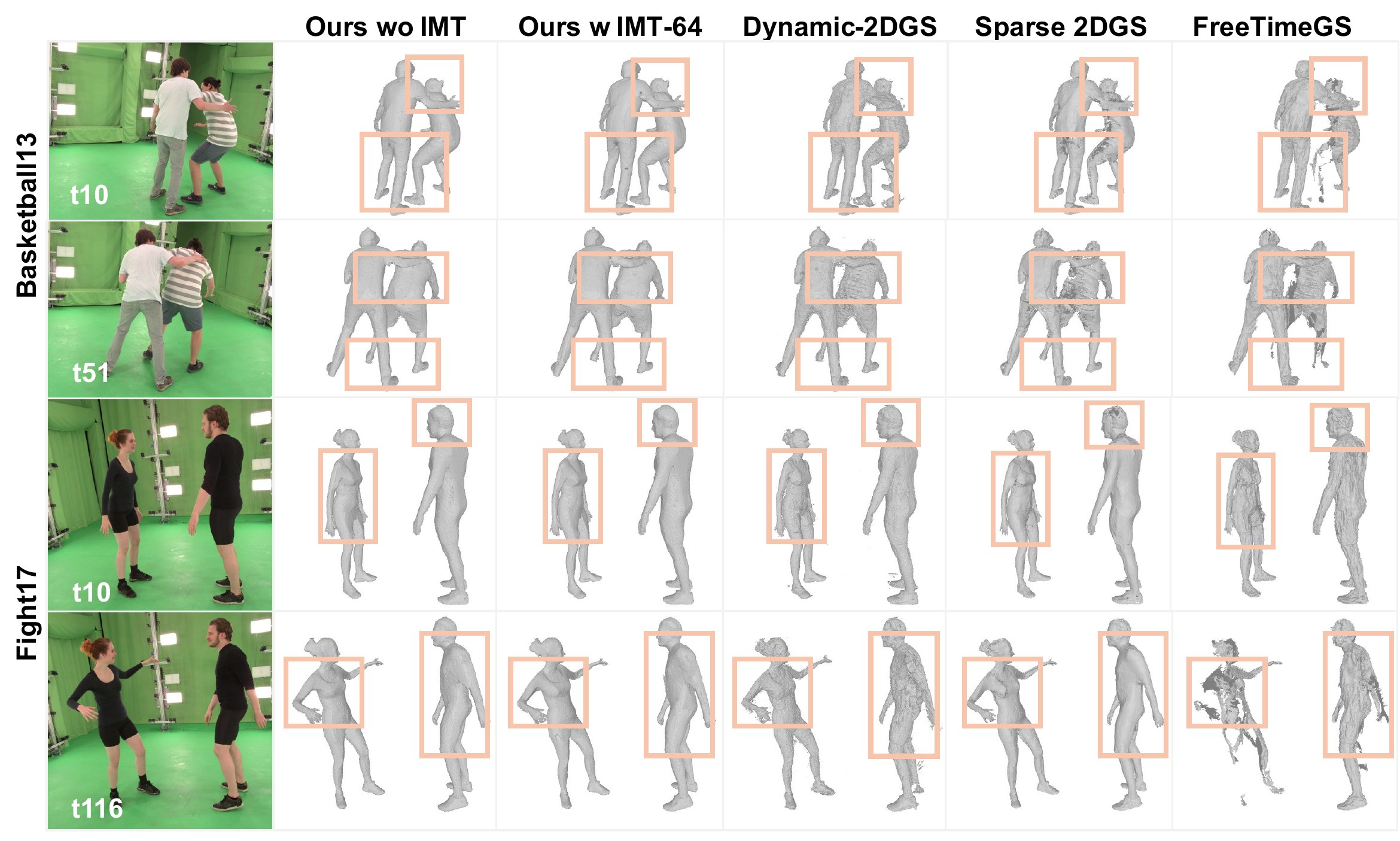}
    \caption{Qualitative results on Hi4D~\cite{yin2023hi4d}. We compare our methods with three baselines (Dynamic-2DGS~\cite{zhang2024dynamic}, Sparse2DGS~\cite{wu2025sparse2dgs}, FreeTimeGS~\cite{wang2025freetimegs}) at two timesteps of the \texttt{Basketball13} and \texttt{Fight17} scene. Bounding boxes highlight major differences.
    }
    \label{fig:hi4d}
\end{figure*}

\section{Experiments}
%%%%%%%%%%%%%%%%%%%%%%%%%%%%%%%%%%%%%%%
\subsection{Settings}

\noindent\textbf{Datasets.}
%%%%%
We conduct experiments on two public datasets: CMU Panoptic~\cite{Joo_2017_TPAMI} and Hi4D~\cite{yin2023hi4d}. Following~\cite{mao2024neural}, we use four scenes from the CMU Panoptic: \texttt{Ian3}, \texttt{Haggling b2}, \texttt{Band1}, and \texttt{Pizza1}, each captured with a circular rig of 10 RGB-D cameras at 1920 $\times$ 1080 resolution. Each scene spans 24 timesteps and provides ground-truth point clouds.
From the Hi4D, we select six scenes: \texttt{Backhug02}, \texttt{Basketall13}, \texttt{Fight17}, \texttt{Football18}, \texttt{Talk22}, and \texttt{Cheers37}, each is captured with 8 RGB cameras at 940 $\times$ 1280 resolution. On average, each sequence contains 118 timesteps and each timestep is annotated with a high-quality textured 3D mesh. Compared with CMU Panoptic, Hi4D features larger motions, longer sequences, and multi-human scene.
Since our goal is surface reconstruction, we use all RGB views for training and only evaluate the meshes.

\smallskip
\noindent\textbf{Evaluation Metrics.}
We follow prior works~\cite{mao2024neural, wang2024space, wu2025sparse2dgs} and adopt Chamfer Distance (CD) that measured in terms of Accuracy $\downarrow$ (Acc), Completeness $\downarrow$ (Comp), and Overall $\downarrow$ (the average of Acc and Comp) as our comparison metrics.

\smallskip
\label{sec:baselines}
\noindent\textbf{Baselines.}~We focus on generic dynamic scene surface reconstruction without relying on external priors, with the emphasis on GS-based methods.
We mainly compare our method against GS-based dynamic surface reconstruction methods, including Space-Time-2DGS~\cite{wang2024space}, Dynamic-2DGS~\cite{zhang2024dynamic}, and GauSTAR~\cite{zheng2025gaustar}.
Note that GauSTAR reconstructs dynamic scenes from dense-view videos using optical flow and depth priors, and performs incremental learning at each timestep.
For completeness, we also compare some NeRF-based dynamic surface reconstruction methods on CMU Panoptic dataset, they are NDR~\cite{cai2022neural} and Neural SDF Flow~\cite{mao2024neural}, as they require long training time.
Some dynamic novel view synthesis methods are also included, such as Tensor4D~\cite{shao2023tensor4d}, 4DGS~\cite{Wu_2024_CVPR}, SC-GS~\cite{huang2024sc}, and FreeTimeGS~\cite{wang2025freetimegs}, which demonstrate strong novel-view synthesis performance, with FreeTimeGS representing the state-of-the-art in novel view synthesis from multi-view dynamic scenes.
We further consider Sparse2DGS~\cite{wu2025sparse2dgs}, a state-of-the-art static surface reconstruction method from sparse views, which we apply independently to each timestep for fair comparison.~We exclude Space-Time-2DGS on the Hi4D due to unavailable code, and also omit MonoFusion~\cite{wang2025monofusion}, DG-Mesh~\cite{liudynamic}, and MaGS~\cite{ma2024mags} for their reliance on strong priors or task-specific assumptions.

\smallskip
\noindent\textbf{Implementation Details.}
We build our model upon~\cite{zhang2024dynamic} and use same parameters for the optimization of Gaussians. 
The initial point cloud is obtained by constructing a visual hull~\cite{yang2024gaussianobject} from all-view foreground masks at the first timestep. 
The segment size is set to 5 with one virtual timestep, and each segment is trained for 30K iterations (about 30 mins). 
The network structure of Gaussian Velocity Field is similar to the deformation network in~\cite{yang2024deformable}. 
If with IMT-64, only the three heads are trainable, while other linear layers are fine-tuned using LoRA (rank=64). 
All experiments are conducted on one NVIDIA RTX 3090Ti GPU. 
For mesh extraction, a TSDF volume~\cite{curless1996volumetric} is constructed by fusing RGB-D images from all training views following~\cite{zhang2024dynamic, Huang2DGS2024}.
More details of datasets, baselines and implementation are in Supplementary Material~Sec. 10.

%%%%%%%%%%%%%%%%%%%%%%%%%%%%%%%%%%%%%%%

\subsection{Comparisons}
We compare our methods against recent GS-based dynamic surface reconstruction baselines on CMU Panoptic~\cite{Joo_2017_TPAMI} in \cref{tab:cmu} and Hi4D~\cite{yin2023hi4d} in \cref{tab:hi4d}. 
We consider two versions of our method: one learns Gaussian Velocity Fields from scratch for each segment (Ours wo IMT), while another version (Ours w IMT-64) employs IMT-64 by fine tuning the Gaussian Velocity Fields.
We present qualitative comparisons on both CMU in \cref{fig:cmu} and Hi4D in \cref{fig:hi4d}.
Baselines used for qualitative comparisons are selected from three categories introduced in \cref{sec:baselines}, with one representative baseline chosen from each category, they are: Dynamic-2DGS~\cite{zhang2024dynamic}, Sparse 2DGS~\cite{wu2025sparse2dgs}, and FreeTimeGS~\cite{wang2025freetimegs}.

\noindent\textbf{CMU Panoptic Dataset.} As shown in \cref{tab:cmu}, our two methods consistently outperforms all the baselines and achieves the lowest Overall values across all four scenes.
Compared with second-best results, Ours w/o IMT surpasses them by over 19\%. 
We visualize the scene \texttt{Band1} and \texttt{Ian3} in \cref{fig:cmu}. 
Since \texttt{Band1} involves a more complex scene, we show two different viewpoints at two distinct timesteps. 
Our methods consistently produce smoother and more detailed reconstructions, whereas the baselines show uneven, noisy, and incomplete surfaces as highlighted in~\cref{fig:cmu}.

\noindent\textbf{Hi4D Dataset.} 
As shown in \cref{tab:hi4d}, our two methods also achieve lowest Overall values on all six scenes.
Ours wo IMT improves upon second-best results by over 49\%.
We visualize reconstructed scenes \texttt{Basketball13} and \texttt{Fight17} at two long-range timesteps from the same viewpoint in \cref{fig:hi4d}. 
It can be clearly observed that our methods reconstruct smoother surfaces and more complete meshes than other baselines. 
Last but not least, both Dynamic-2DGS and FreeTimeGS exhibit severe geometric inconsistencies and accumulated errors over timesteps. 
Also, their reconstructed surfaces become noticeably coarse and jittering. 
While our methods maintain best geometric consistency and smooth surface across long-range timesteps.

%%%%%%%%%%%%%%%%%%%%%%%%%%%%%%%%%%%%%%%

% \usepackage{graphicx}
\begin{table}[t]
\centering
\caption{Ablation Studies on the Hi4D dataset~\cite{yin2023hi4d}. 
We calculate the average of the three metrics (Acc, Comp, and Overall) for the six scenes.
GVF: Gaussian Velocity Field. 
SF-Reg: SDF-Flow Regularization. 
I-Segment: Independent Segment Partitioning. 
O-Segment: Overlapping Segment Partitioning. 
IMT-64: Incremental Motion Tuning with LoRA rank 64.}
\label{tab:ablation}
\resizebox{\columnwidth}{!}{%
\begin{tabular}{ccccccccc}
\hline
& GVF      & SF-Reg  & I-Segment & O-Segment & IMT-64    & Acc $\downarrow$ & Comp $\downarrow$ & Overall $\downarrow$ \\ \hline
\textbf{(a)} & $\surd$ &         &         &         &         & 1.75              & 1.23               & 1.49                  \\
\textbf{(b)} & $\surd$ & $\surd$ &         &         &         & 1.18              & 0.86               & 1.02                  \\
\textbf{(c)} & $\surd$ & $\surd$ & $\surd$ &         &         & 1.06              & 0.47               & 0.77                  \\
\textbf{(d)} & $\surd$ & $\surd$ &         & $\surd$ &         & 0.89             & 0.45              & 0.67                 \\
\textbf{(e)} & $\surd$ & $\surd$ &         & $\surd$ & $\surd$ & 0.91             & 0.48              & 0.70                 \\ \hline
\end{tabular}%
}
\end{table}

\begin{figure}[t]
    \centering
    \includegraphics[width=\columnwidth]{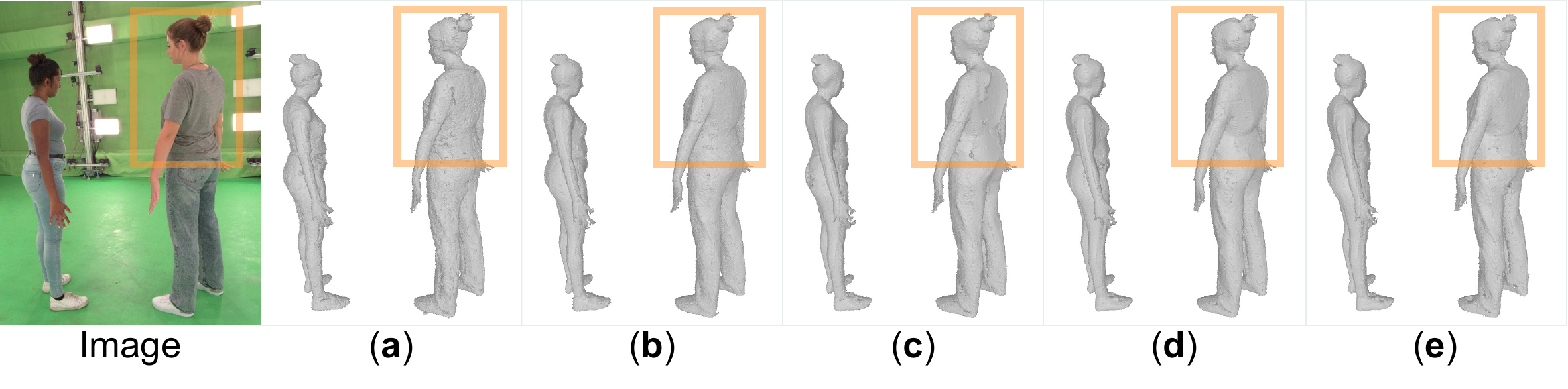}
    \caption{
    Qualitative comparison of each ablation study on scene \texttt{Cheers37}. 
    Subscripts correspond to rows in \cref{tab:ablation}.
    (\textbf{a}): Gaussian Velocity Field with normal \& depth regularization from ~\cite{Huang2DGS2024}, photorealistic and mask losses.
    (\textbf{b}): Add SDF Flow Regularization based on (\textbf{a}).
    (\textbf{c}): Training (\textbf{b}) with Independent Segment Partitioning.
    (\textbf{d}): Training (\textbf{b}) with Overlapping Segment Partitioning.
    (\textbf{e}): Adopting Incremental Motion Tuning (LoRA rank 64) on (\textbf{d}).
    }
    \label{fig:viz-sdf-flow}
\end{figure}

% \usepackage{graphicx}
% \vspace{-1em}
\begin{table}[t]
\centering
\caption{Temporal stability comparison on the Hi4D dataset~\cite{yin2023hi4d}. STD: standard deviation. It shows the average STD of the three metrics (Acc, Comp, and Overall) across six scenes. The best and the second-best are highlighted in \textbf{bold} and \underline{underlined}.}
\label{tab:std}
\resizebox{0.75\columnwidth}{!}{%
\begin{tabular}{cccc}
\hline
Methods                              & Acc STD $\downarrow$ & Comp STD $\downarrow$ & Overall STD $\downarrow$ \\ \hline
Dynamic-2DGS~\cite{zhang2024dynamic} & 1.50                 & 1.30                  & 1.19                     \\
GauSTAR~\cite{zheng2025gaustar}      & 0.53                 & 4.39                  & 2.70                     \\
Sparse2DGS~\cite{wu2025sparse2dgs}   & 0.95                 & 0.50                  & 0.68                     \\ \hline
Ours w IMT-64                     & \underline{0.28}               & \underline{0.37}                 & \underline{0.28}                     \\
Ours wo IMT                                 &{\bf 0.22}                 & {\bf 0.17}                  & {\bf 0.18}                    \\ \hline
\end{tabular}%
}
\end{table}

% \usepackage{graphicx}
\begin{table}[t]
\centering
\caption{Different LoRA ranks comparison on Hi4D dataset~\cite{yin2023hi4d}. It shows the average of the three metrics (Acc, Comp, and Overall) in six scenes under different LoRA ranks.}
\label{tab:rank}
\resizebox{0.55\columnwidth}{!}{%
\begin{tabular}{cccc}
\hline
Method & Acc $\downarrow$ & Comp $\downarrow$ & Overall $\downarrow$ \\ \hline
Ours w IMT-16 & 1.07      & 0.70          & 0.89     \\
Ours w IMT-32 & 1.00      & 0.64          & 0.82     \\
Ours w IMT-64 & 0.91     & 0.48         & 0.70    \\
Ours wo IMT   & 0.89     & 0.45         & 0.67    \\ \hline
\end{tabular}%
}
\end{table}

\begin{figure}[t]
    \centering
    \includegraphics[width=0.95\columnwidth]{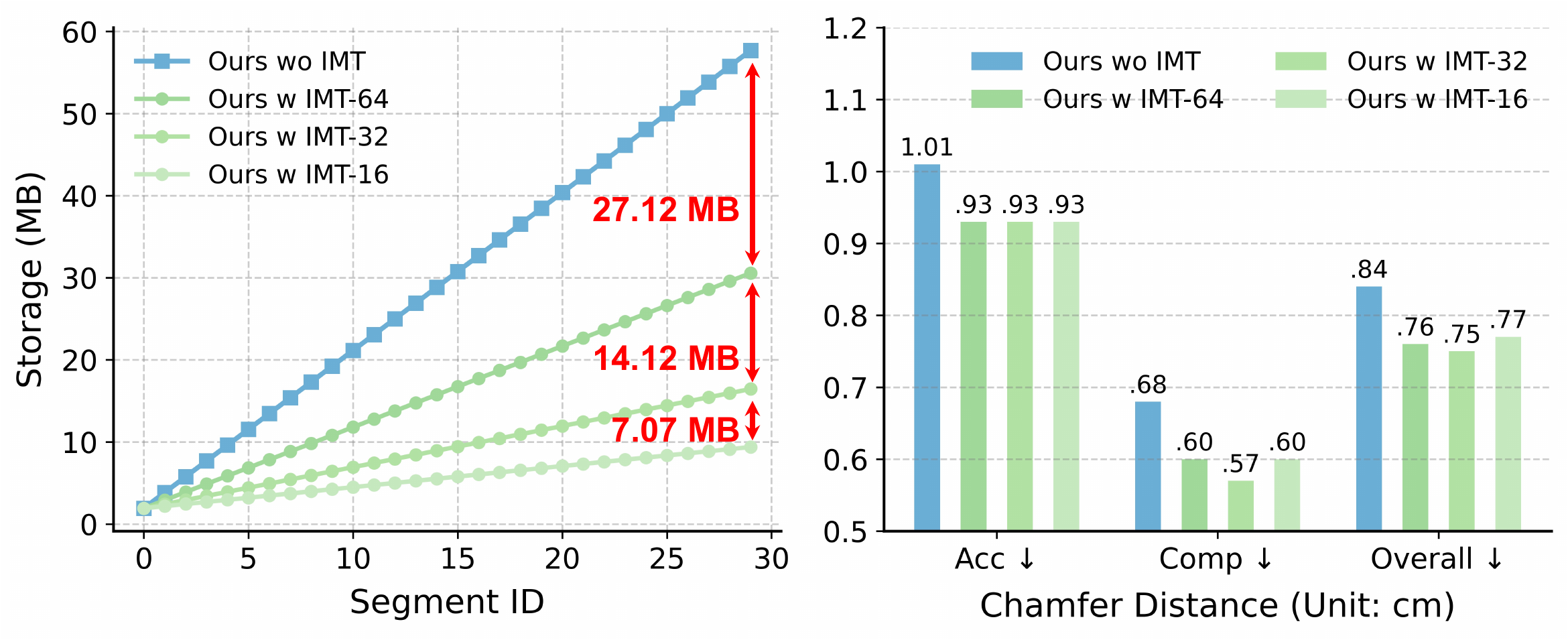}
    \caption{
    LoRA rank and storage analysis on scene \texttt{Backhug02}.
    \textit{Left}: As the number of segments increases, \textit{Ours wo IMT} exhibits growing storage of Gaussian velocity fields, while \textit{Ours w IMT-\{LoRA-rank\}} effectively curb the storage growth.
    \textit{Right}: For different LoRA ranks, \textit{Ours w IMT} maintains strong performance (even at rank 16), achieving competitive results to \textit{Ours wo IMT}.
    }
    \label{fig:rank}
\end{figure}

\subsection{Ablation Studies \& Discussion}
We perform ablation studies and discussion on Hi4D~\cite{yin2023hi4d} because it has longer sequences with larger deformations.

\noindent\textbf{Ablation Studies.} We conduct ablation studies on each component and present quantitative and corresponding qualitative results in \cref{tab:ablation} and \cref{fig:viz-sdf-flow}.
We first replace the deformation field in Dynamic-2DGS~\cite{zhang2024dynamic} with our proposed \textit{Gaussian Velocity Field} while keeping the depth and normal regularization, photorealistic and mask losses unchanged (row~\textbf{a}, \cref{fig:viz-sdf-flow}(\textbf{a})). 
Subsequently, we incorporate the \textit{SDF Flow Regularization} (row~\textbf{b}, \cref{fig:viz-sdf-flow}(\textbf{b})) to encourage consistent surface evolution from motion of Gaussians and geometry changes across time, which leads to a noticeable improvement in reconstruction performance. As shown in \cref{fig:viz-sdf-flow}(\textbf{b}), using regularization can be much smoother than not using this regularization term and artifacts are reduced.
We then apply \textit{Independent Segment Partitioning}, dividing the sequence into independent segments of size 5 (row~\textbf{c}, \cref{fig:viz-sdf-flow}(\textbf{c})). This design effectively reduces error accumulation in both Acc, Comp and Overall.
However, lacking geometric information passing across segments, we introduce the \textit{Overlapping Segment Partitioning} strategy (row~\textbf{d}, \cref{fig:viz-sdf-flow}(\textbf{d})), which further improves performance.
Comparing \cref{fig:viz-sdf-flow}(\textbf{c}) and (\textbf{d}), it is easy to see that with the Gaussians of the previous segment's virtual timestep as the initialization of the next segment, the geometric quality and consistency are improved, and the surface becomes smoother.
Finally, adopting \textit{IMT-64} (row~\textbf{e}, \cref{fig:viz-sdf-flow}(\textbf{e})) achieves comparable performance quantitatively and qualitative while reducing storage.
Below, we compare the stability of our method with that of other baselines and the performance of IMT with varying ranks. Additionally, please see Supplementary Material for more experiments.

\smallskip
\noindent\textbf{Temporal Stability.} 
We evaluate the temporal stability using the average standard deviation (STD) of three metrics on six scenes of Hi4D dataset~\cite{yin2023hi4d}.
\cref{tab:std} shows that our methods achieve the lowest average STD across all metrics, demonstrating superior temporal stability compared with other baselines.
Specifically, our method without IMT-64 reduces the STD of CD (Overall metric) to only $0.18$, significantly lower than Sparse2DGS~\cite{wu2025sparse2dgs} ($0.68$) and Dynamic-2DGS~\cite{zhang2024dynamic} ($1.19$). This indicates that the reconstructed surfaces from our methods exhibit substantially fewer surface jitter and are more temporally consistent.

\noindent\textbf{LoRA Rank \& Storage Analysis.}
We try different LoRA ranks (16, 32, 64) on fine-tuning Gaussian Velocity Fields using IMT, excluding rank~128 as its parameters approach full fine-tuning. 
Results in~\cref{tab:rank} show that lower ranks slightly reduce performance due to limited capacity but still maintain competitive results. 
Take the scene \texttt{Backhug02} as an example. While canonical shape storage per segment remains stable on all settings (38–39~MB), \cref{fig:rank}(left) shows all Gaussian Velocity Fields' storage drops from 57.7~MB (Ours wo IMT) to 30.6, 16.5, and 9.4~MB for IMT-64, IMT-32, and IMT-16, respectively. \cref{fig:rank}(right) demonstrates that IMT maintains well reconstruction quality with different ranks, making IMT scalable for long sequences.

\section{Conclusion}
In this paper, we introduce a novel method \textit{``4DSurf''} for generic dynamic scene surface reconstruction from sparse view videos. 
Our key idea is to regularize geometry evolution by matching the SDF flow from motion of Gaussians and geometry changes, which enforces temporally consistent surface evolution. 
To further alleviate deformation error accumulation and storage usage, we propose Overlapping Segment Partitioning and Incremental Motion Tuning, respectively. 
Extensive experiments on two challenging datasets demonstrate that our method achieves the lowest and temporally-stablest in Chamfer distance.

\noindent\textbf{Acknowledgement. }~ML is supported partially by an ARC Discovery Grant: DP200102274. 
HL holds concurrent appointments with both ANU and Amazon.   This paper describes work performed at ANU and is not associated with Amazon. HL is also partially supported by an ARC Discovery Grant: DP220100800.

{
    \small
    \bibliographystyle{ieeenat_fullname}
    \bibliography{main}
}

\end{document}